# Q-Learning with Basic Emotions

Wilfredo Badoy Jr.
Department of Information System and Computer Science
Ateneo de Manila University
Quezon City, Philippines
wbadoy@yahoo.com

Kardi Teknomo
Department of Information System and Computer Science
Ateneo de Manila University
Quezon City, Philippines
teknomo@gmail.com

*Abstract*—Q-learning is a simple and powerful tool in solving dynamic problems where environments are unknown. It uses a balance of exploration and exploitation to find an optimal solution to the problem. In this paper, we propose using four basic emotions: joy, sadness, fear, and anger to influence a Q-learning agent. Simulations show that the proposed affective agent requires lesser number of steps to find the optimal path. We found when affective agent finds the optimal path, the ratio between exploration to exploitation gradually decreases, indicating lower total step count in the long run.

## I. INTRODUCTION

Imagine a world where humans and robots are indistinguishable from each other. Where robots interact with us like normal people do. Wouldn't it be nice if they can feel our emotions, and in turn they can exhibit emotions of their own? Although that is very far in the future, steps have been made in to that direction. Several researches have been done combining computer learning with affect even in the early 80's. These experiments where usually set in a controlled physical environment with a mechanical robot that is fed with goals and moved through the environment using some sort of reinforcement learning procedure. These robots can either accept additional rewards through human intervention or from their own actions. Today, similar experiments can been done on discrete environments with artificial virtual agents.

Robots are commonplace today in industries, they assist humans with work that humans could never do physically, such as lifting heavy parts or working in hostile environments. As these robots will leave the realm of the industry into our homes and workplaces, it is important that we interact with them efficiently and that is comfortable for both humans and agents. Humans not just interact with information but also with emotion. In fact, negative emotion enhances memory accuracy [1] and positive emotion broaden our scope of attention [2].

Different affect models have been used in the past to incorporate some sort of emotion in the learning process of robots and agents. Arousal and pleasure factors have been used to influence and agent's movement.

Our paper incorporates higher level emotions such as fear, anger, sadness, and joy in agent learning, specifically Q-learning partially based on Korsten and Taylor's model [5]. Our main purpose is to investigate the difference or similarity in the number of exploration steps between a normal q-learning agent versus an agent whose decision is based on circumstances where it can mimic joy, sadness, fear, and/or anger. Two main performance indicators will be explored: the average number of steps per episode, and the average number of steps until the optimal path is found.

Equation 1 is the Temporal Difference update equation for Q-learning

$$Q_{t+1}(s_t, a_t) \leftarrow Q(s_t, a_t) + \alpha [r_{t+1} + \gamma \max_a Q(s_{t+1}, a) - Q(s_t, a_t)] \quad (1)$$

where $s_t$ is the current state, and $a_t$ is the action taken; $a_t$ is also the next state $s_{t+1}$. A new quality value $Q_{t+1}(s_t, a_t)$ is calculated from the old value $Q(s_t, a_t)$ and a correction. The correction is based on learning rate $\alpha$ which determines to what extent the newly acquired information will override the old information, the reward value $r_{t+1}$ which is observed after performing $a_t$ in $s_t$, the discount factor $\gamma$ which determines the importance of future rewards, and $\max_a Q(s_{t+1}, a)$ the maximum $Q$ for all state-action pairs of the next state.

We used the $\varepsilon$-greedy algorithm as stated in [6] for its action selection policy. In this policy, most of the time they choose an action that has maximal estimated reward value, but with a probability $\varepsilon$, they instead select an action at random. Learning rate $\alpha$ and discount rate $\gamma$ will be held at a constant rate, while only the $\varepsilon$-greedy parameter $\varepsilon$ will be made to vary for investigation.

Since $\varepsilon$ does not govern the action of the proposed agent, it will be the only parameter that should differ in the number of steps taken by the normal agent and the proposed agent. The number of steps taken by the proposed agent is unaffected by changes in $\varepsilon$, while the normal agent's number of steps vary with the change in $\varepsilon$. Only four basic emotions such as joy, sadness, fear, and anger will be considered. Simulation is limited to one agent (predator) pursuing a static goal (prey).

In this work, a novel approach to incorporate emotion into Q-learning is developed. This work is novel on two accounts. First, this is the first time basic emotions have been incorporated into Q-learning. Second, this is the first time basic emotions have been made to influence and agent's behavior in terms of direction and speed.

## II. REVIEW OF RELATED LITERATURE

To effectively explore the idea of agent learning with the influence of emotions, a subset of literature has been selected to answer the following questions:

A. How emotion was modeled and how did these models influenced agent learning?
B. How the experiments were done and what are the results?

### A. How emotion was modeled and how did these models influenced agent learning?

In the work of El-Nasr et al [7], [8], [9], user feedback is used to provide action values. Broekens [10] used human's emotional expressions. These expressions are analyzed in real time and converted to an additional reinforcement signal used by the robot; positive expressions result in reward, negative expressions in punishment. They took affect to mean the positiveness versus the negativeness (valence) of a situation, object, etc. User feedback is a form of extrinsic motivation. Most of the researches though are focused on intrinsic motivation.

Intrinsic means dealing with internal reward rather than external input. Gadanho et al. [12-15] used a complex model that receives input from the environment by sensors. This input is processed through a Feelings module, which outputs feelings such as Hunger, Pain, Restlessness, Temperature, Eating, Smell, Warmth and Proximity. This in turn serves as input to an Emotion module, which outputs Joy, Sadness, Fear, and Anger. Some component of these emotions is fed back into the feelings module through the Hormone component to include the suggestion of [16]. Only one dominant emotion is selected, which can either be a good or bad emotion. A value from this observation is then fed into the learning function. This learning function then chooses between three behaviors, namely; avoid-obstacles, seek-light, and wall-follow. In a previous work, the researchers proposed to use a model with a goal system [17]. Emotions are not modelled explicitly by the goal system, although it is often inspired by them. The goal system is built on a set of homoeostatic variables that it tries to maintain within fixed bounds. The output of this goal system provides a reinforcement reward for the learning algorithm.

Ahn and Picard [18] proposed a system where motivation in learning and decision making are influenced by cognition and emotion as internal rewards and external rewards from the environment. Two matrices represented the Q for cognition and emotion, and the internal value is a function of these two. The same researcher also proposed an affect model based on valence and arousal. The valence is positive when the current choice is expected to give a higher than average reward, and negative otherwise. Arousal is a function of uncertainty. Valence and arousal define the intrinsic reward of the Q-learning algorithm [19].

Marinier and Laird [20] set up an environment that was characterized into high/low suddenness, high/low pleasantness, high/low relevance, and high/low conductiveness. Based on these characteristics, an appraisal mechanism provides a valenced feeling intensity which serves as a reward signal over which the agent learns. Sequeira et al. [21] proposed to use novelty, motivation, valence and control as appraisal dimensions. These appraisals were turned into scalars, and an intrinsic reward is calculated as a linear combination of these values. More recently, Kuremoto et al. used a simplified circumplex model of affect by Russell [4]. They proposed adopting Arousal and Pleasure into conventional Reinforcement Learning to improve the performance of Reinforcement Learning in multi-agent systems. They suggested that the fundamental affect factors Arousal and Pleasure be used to produce an emotion function, and the emotion function is combined with Q function which is the state-action value function in Q-learning to constitute a motivation function. The motivation function is implemented into the stochastic policy function of Reinforcement Learning instead of Q function in QL. Agents select behaviors not only according to the states they observed from the environment, but also to their internal affective responses [4].

### B. How the experiments were done and what are the results?

Experiments were either done using a physical autonomous robot or a virtual autonomous agent. Two notable robots are Khepera [11] and RoCo [18]. Khepera is the preferred experimental robot by Gadanho and Hallam [13], [14]. The Khepera robot is a small robot with two wheels, and eight infrared sensors that allow it to sense object proximity and light. Sensors are located in both the front and rear of the robot. The robot's task is to move around the environment, find different food sources, and extract energy for the food source.

They [12], [13] made four identical experiments using different controllers: hand-crafted, event-triggered, interval-triggered, and random. The controller is responsible for making the robot move. In these experiments, they showed that emotions can be used as both reinforcement and event detector. The emotion-dependent event detector allows drastic cuts in the number of triggering of the learning controller. It learns its task with much less iterations and needs lesser control. In another study [16], they have observed that the emotional controller has lesser number of events and collisions. They conclude that while emotional association may be powerful in their ability to cover states, they don't have the explanatory power and may introduce errors of over-generalization.

Cognitive power on the other hand, is more accurate but only selects the few instances that seem most important.

Ahn and Picard used a physical autonomous robot named RoCo. RoCo is a collaboration of MIT Media Lab's Affective Computing Group and Robotic Life Group. RoCo's goal is to try and solicit attention and pleasure from the user [18], [19].

El-Nasr, et al. [7], [8], [9] used a simulated pet named PETEEI. PETEEI is acronym for a PET with Evolving Emotional Intelligence. Users interact with the pet with pre-defined actions, and the pet provides a feedback from a pre-defined set of sounds. The user is then provided with a survey form to rate the believability of the pet. They concluded that emotion improved the believability of the pet. In a separate study [9], they found out that the introduction of learning improved the system and created a closer simulation of the behavior of a real pet.

In a study by Broekens [10], a simulated robot (agent) is set up to live in a continuous grid world environment consisting of wall, food and path patches. These are the features of the world observable by the agent. The agent cannot walk on walls, but can walk on path and food. Walls and path are neutral (have a reinforcement of 0.0), while food has a reinforcement of 10. One cell in the grid is assumed to be a 20 by 20 object. An experiment was then made to compare social reinforcement learning with the normal" on-social" reinforcement learning. In this experiment he showed that affective interaction in human-in-the-loop learning can provide a significant benefit to the efficiency of a reinforcement learning robot in a continuous grid world.

Ahn and Picard [19] simulated both single-step decision making with a two-armed bandit type gambling task and a maze task. They have showed that addition of affective anticipatory reward can be used for improving the effectiveness of learning and decision making. In a study by Marinier and Laird [20], a maze was made for the agent to navigate. Three agent types were tested: a standard reinforcement learning agent, an agent with emotions but not mood, and a full agent that included mood. They showed that agents with emotion learn very fast relative to standard reinforcement learning agent.

Kuremoto et al. [4] made computer simulations of pursuit problems to verify the effectiveness of the proposed method. An environment with 17x17 grids is set up. Prey does not move at position (9, 9), while two hunters (o) find and capture the prey. One hunter starts from (2, 2) and while another starts at (14, 14). Experiments were then made to compare the proposed method and plain Q-learning method. Results have shown that learning efficiency was enhanced in the proposed method as compared to conventional Q-Learning.

## III. CONCEPTUAL FRAMEWORK

Fig. 1 shows the conceptual framework of the proposed agent. The explanation starts with the bottommost square. The position of the agent relative to the target gives rise to certain emotions (states). Table I shows the value comparisons and the resulting emotion. Act represents the actual value, norm represents the average, and exp1 represents the expected value based on the current stimuli. The proposed agent will only use a subset of what is shown in Table I. It will only include four basic emotions, namely: joy, sadness, fear, and anger. Although Ekman [22] listed six which includes disgust and surprise, it can be argued that disgust is a drive rather than an emotion, as it is strongly related to body state, similarly to other drives like pain and hunger [5]. Jack et al [23] also shows that humans only have four basic emotions, which does not include surprise and disgust.

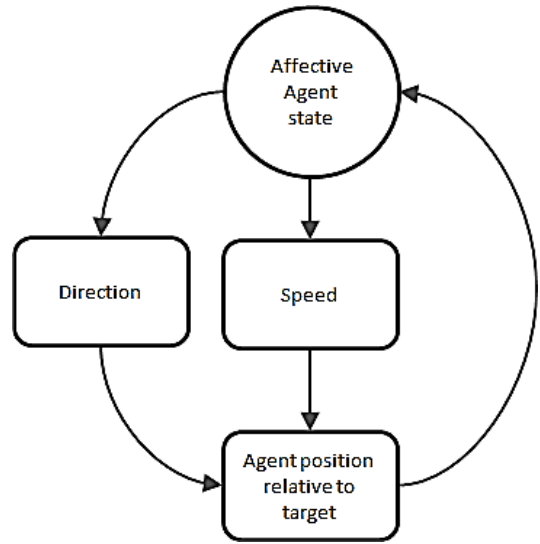

Fig. 1. Conceptual Framework

TABLE I. VALUE COMPARISONS FOR THE REWARD/PUNISHMENT VALUE TYPE [5]

| Value comparison | Reward/Punishment |
|---|---|
| exp1 <norm | fear |
| exp1 >norm hope | hope |
| act <norm & exp1 >act | anger |
| act <norm & exp1 =<act | sadness |
| exp2 <norm & exp1 >exp2 | relief |
| exp2 >norm & exp1 <exp2 | disappointment |
| act >norm & exp2 <act | surprise |
| act >norm | joy |

Table II shows the adapted value comparisons used in this study. Average value here is estimated from a regression line using the current and previous number of steps. Power regression, also known as the log-log regression takes the input signal and fits a function $y = at^b$ to it where $t$ is the variable along the $x$-axis. The function is based on the function linear regression, with both axes scaled logarithmically [24]. The power regression line shows the best fit so far. The expected number of steps is initially equal to the average value, and increments every step past the average. The actual value is the shortest known path from the current position to the target. In the event of an unknown position, it maintains the value of the previous actual path.

TABLE II. REVISED VALUE COMPARISONS AND ASSOCIATED EMOTIONS

|  | Reward/Punishment |
|---|---|
| exp1 <norm | fear |
| act <norm & exp1 >act | anger |
| act <norm & exp1 =<act | sadness |
| act >norm | joy |

Study by [25] participants with normal emotion processing were engaged in a card-drawing task. When drawing from "dangerous decks" and consequently experiencing losses and the associated negative emotions, they subsequently made safer and more lucrative choices. In other words, after experiencing negative affect such as sadness, anger, and fear, humans tend to prefer safer and higher rewards choices. Sad individuals are motivated by an implicit goal of reward acquisition [26]. Angry people tend to process heuristically, not thinking of alternate solutions [27]. On the other hand, happy individuals are more likely to optimize or sacrifice in their decision making, rather than maximizing to achieve the best outcome [28].

Emotions were found to strongly affect the kinematics of locomotion, particularly walking speed. Barliya et al. [28] found out the anger and happiness being more "energetic", than sadness and fear, and also in relation to speed. This observation also mirrors the study of Crane and Gross [29] which says happy and angry people tend to move faster than sad and frightened ones. Table III summarizes the effect of these emotions on speed.

TABLE III. EMOTION TRANSLATED TO AGENT DECISION AND SPEED

| Emotion | Direction | Speed |
|---|---|---|
| Joy | random | fast |
| Sadness | greatest reward | slow |
| Anger | greatest reward | fast |
| fear | greatest reward | slow |

IV. METHODOLOGY

To compare the performance of a standard Q-learning agent and the proposed affective agent, simulations of pursuit problems will be executed. A discrete environment is set up consisting of a 15x15 grid. The goal or prey is situated in the center at position (6, 6). The autonomous agent is set at position (0, 0) at the start. This setup is an adaptation of Kuremoto's test [4]. Fig. 2 shows the test environment.

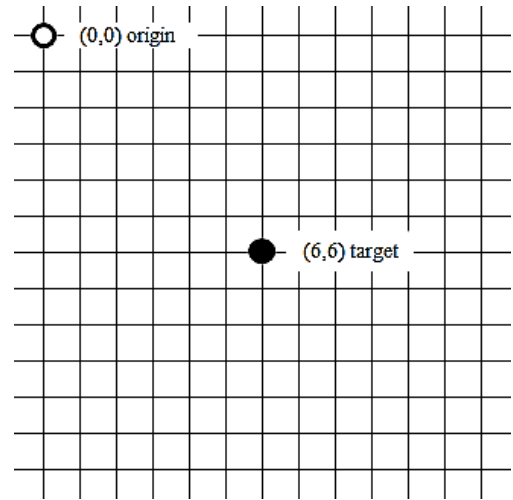

Fig. 2. Test environment

Each agent will be go through 20 simulations with 200 episodes each, for each value of $\varepsilon$. Parameter $\varepsilon$ will be varied from 0.1 to 0.9 in increments of 0.1 steps. The number of steps taken is measured per episode taking note of the episode where the optimal path to the target is found. The standard agent will be using $\varepsilon$-greedy action selection algorithm to select the next position, whether that be up, down, left, or right of the current position. For the proposed agent, the decision to select the next position depends on whether the agent is happy, in which case it randomly chooses its next position and goes for the highest reward otherwise. The standard agent will move at a constant one step at-a-time fashion, while the proposed agent will either move 0 or 1 step if it is afraid or sad, and 1 or 2 steps if it is angry or happy. The decision to which number of steps to take will be chosen at random.

V. ANALYSIS

The performance of both standard Q-learning agent and the proposed affective agent is compared.

Table IV doesn't show significant difference between the two agents in terms of number of steps per episode but Fig. 3 shows that for epsilon below 0.5, the standard agent has lower step count per episode, and above 0.6, the proposed agent performs better. Because the proposed agent doesn't take into account the value of $\varepsilon$, it shows a constant value across the $\varepsilon$ range.

TABLE IV. T-TEST COMPARISON FOR AVERAGE NUMBER OF STEPS PER EPISODE.

|  | ε-greedy selection | proposed |
|---|---|---|
| Mean | 67.02 | 56.28 |
| Variance | 1626.05 | 9.80 |
| Observations | 9 | 9 |
| Pearson Correlation | 0.36 |  |
| Hypothesized Mean Difference | 0.00 |  |
| df | 8 |  |
| t Stat | 0.82 |  |
| P(T<=t) one-tail | 0.22 |  |
| t Critical one-tail | 1.86 |  |
| P(T<=t) two-tail | 0.44 |  |
| t Critical two-tail | 2.31 |  |

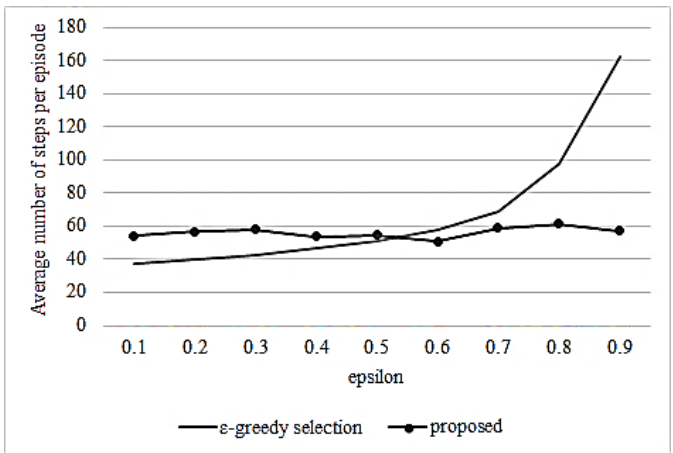

Fig. 3. Average number of steps per episode versus $\varepsilon$

Table V shows the statistical comparison of total number of steps before the optimal path is found. T-test shows that in finding the optimal path, the proposed affective agent performs better. This is evident in Fig. 4 which shows the proposed agent has lesser number of steps except when $\varepsilon = 0.7$.

TABLE V. T-TEST COMPARISON FOR TOTAL NUMBER OF STEPS BEFORE OPTIMAL PATH IS FOUND.

|  | ε-greedy selection | proposed |
|---|---|---|
| Mean | 110190.20 | 99357.56 |
| Variance | 76689863 | 53278480 |
| Observations | 9 | 9 |
| Pearson Correlation | -0.0623 |  |
| Hypothesized Mean Difference | 0 |  |
| df | 8 |  |
| t Stat | 2.77 |  |
| P(T<=t) one-tail | 0.01 |  |
| t Critical one-tail | 1.86 |  |
| P(T<=t) two-tail | 0.02 |  |
| t Critical two-tail | 2.31 |  |

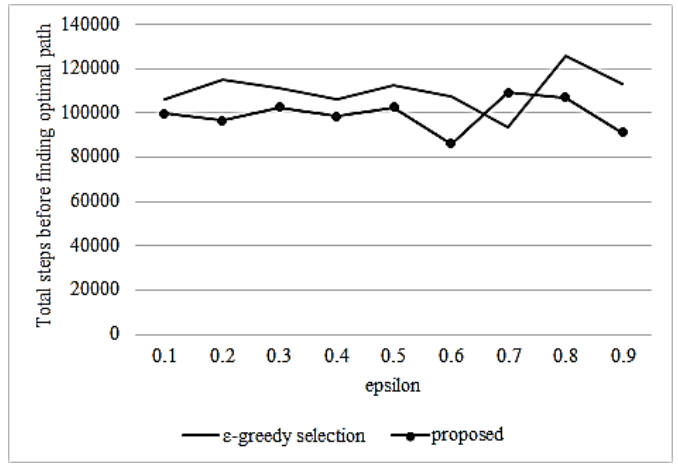

Fig. 4. Total number of steps before optimal path discovery

Analysis of the affective agent's affect profile show that the dominant emotion that comes into play are joy and anger, joy being the dominant emotion in the earlier episodes, which would be superseded by anger in the latter episodes. Fig. 5 shows you this behavior. This means that the proposed agent will explore more in the earlier episodes, and will exploit more in the latter episodes.

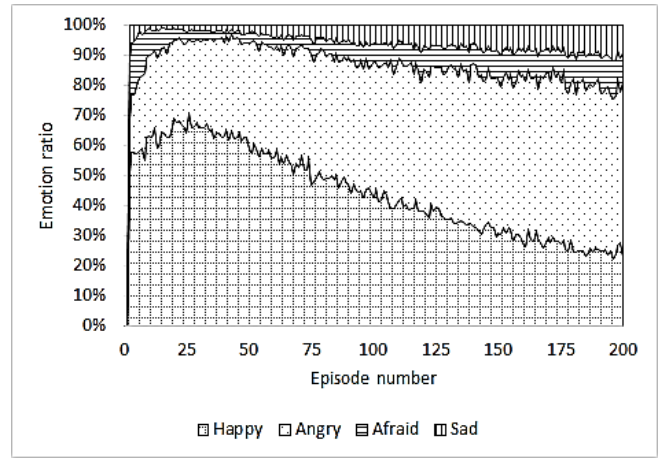

Fig. 5. Affective agent emotion profile

Calculating the equivalent randomness of the steps shows the equivalent $\varepsilon$ profile in Fig. 6. The figure shows a varying $\varepsilon$ throughout the simulation, starting high reaching about 0.3 and decreasing to 0.1 near the end. It is also interesting to note that for the proposed affective agent, the average number of episodes required to find the optimal path is 27.35. This value coincides where the equivalent $\varepsilon$ of the proposed affective agent tapers off in Fig. 6, which seems to suggest that it inherently "knows" that it has already found the optimal path and may not need to randomize so much anymore to find another optimal path.

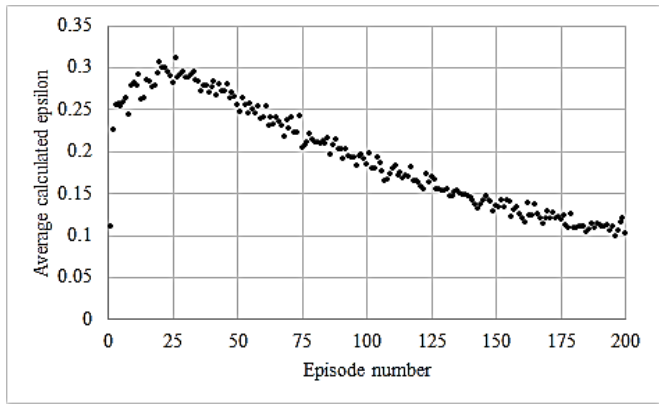

Fig. 6. Equivalent $\varepsilon$ per episode for affective agent

## VI. Conclusion And Recommendations

To investigate the influence of emotions on a Q-learning agent, an affective agent is proposed in this paper. Results show that the proposed agent requires less number of steps in finding the optimal path, and gradually decreases its exploration/exploitation ratio. The gradual decrease in the exploration/exploitation ratio indicates that in the long run, the total number of steps of the affective agent will be lower than that of the standard Q-learning agent.

More study should be done in this area, especially with multiple agents, either in competitive or cooperative mode. Another avenue of further study would be to test this theory on targets that move. And finally, further study should be done in incorporating behavior into this proposed agent, to compare the performance of an optimistic agent to a pessimistic one.

## Acknowledgment

This research is partially supported by the Philippine Council for Industry, Energy and Emerging Technology Research and Development Department of Science and Technology (PCIEERD-DOST) and Philippine Higher Education Research Network (PHERNET).